\documentclass[letterpaper, 10 pt, conference]{ieeeconf}  

\IEEEoverridecommandlockouts                              

\usepackage{graphics} 
\usepackage{epsfig} 
\usepackage{times} 
\usepackage{amsmath} 
\usepackage{amssymb}  
\usepackage{color}
\usepackage{tikz}
\usepackage{subfig}
\usepackage{xfrac}
\usepackage{bm}
\usepackage{cite}
\usepackage[normalem]{ulem}

\title{\LARGE \bf
A Framework for Depth Estimation and Relative Localization \\ of Ground Robots using Computer Vision*
}

\author{R\^{o}mulo T. Rodrigues$^{1}$, Pedro Miraldo$^{2}$, Dimos V. Dimarogonas$^{2}$, and A. Pedro Aguiar$^{1}$
\thanks{*This work was supported in part by SYSTEC UID/EEA/00147/2019 funded by national funds through the FCT/MCTES (PIDDAC); Project STRIDE NORTE-01-0145-FEDER-000033, supported by NORTE 2020, under the Portugal 2020 Partnership Agreement, through the European Regional Development Fund (ERDF); PDMA - NORTE-08-5369-FSE-000061, through NORTE 2020, IMPROVE POCI-01-0145-FEDER-031823, funded by FEDER funds through COMPETE2020, POCI and PIDDAC; COIN from Swedish Foundation for Strategic Research, Knut och Alice Wallenberg Foundation (KAW); and by the EU H2020 Co4Robots project.}
\thanks{$^{1}$R. T. Rodrigues A. P. Aguiar are with the Research Center for Systems and Technologies (SYSTEC), Faculty of Engineering, University of Porto, Porto, Portugal.\newline
        E-Mail:{\tt\small rtr@fc.up.pt} and {\tt\small pedro.aguiar@fe.up.pt}.} %
\thanks{$^{2}$P. Miraldo and D. V. Dimarogonas are with the Devision of Decision and Control Systems, KTH Royal Institute of Technology, Stockholm, Sweden.\newline
        E-Mail:{\tt\small \{miraldo,dimos\}@kth.se}.}%
}

\makeatletter
\renewcommand{\maketag@@@}[1]{\hbox{\m@th\normalsize\normalfont#1}}%
\makeatother

\begin{document}

\maketitle
\thispagestyle{empty}
\pagestyle{empty}

\begin{abstract}
The 3D depth estimation and relative pose estimation problem within a decentralized architecture is a challenging problem that arises in missions that require coordination among multiple vision-controlled robots. The depth estimation problem aims at recovering the 3D information of the environment. The relative localization problem consists of estimating the relative pose between two robots, by sensing each other's pose or sharing information about the perceived environment. Most solutions for these problems use a set of discrete data without taking into account the chronological order of the events. This paper builds on recent results on continuous estimation to propose a framework that estimates the depth and relative pose between two non-holonomic vehicles. The basic idea consists in estimating the depth of the points by explicitly considering the dynamics of the camera mounted on a ground robot, and feeding the estimates of 3D points observed by both cameras in a filter that computes the relative pose between the robots. We evaluate the convergence for a set of simulated scenarios and show experimental results validating the proposed framework.

\end{abstract}

\section{INTRODUCTION}

\begin{figure}
    \centering
    \includegraphics[width=.47\textwidth]{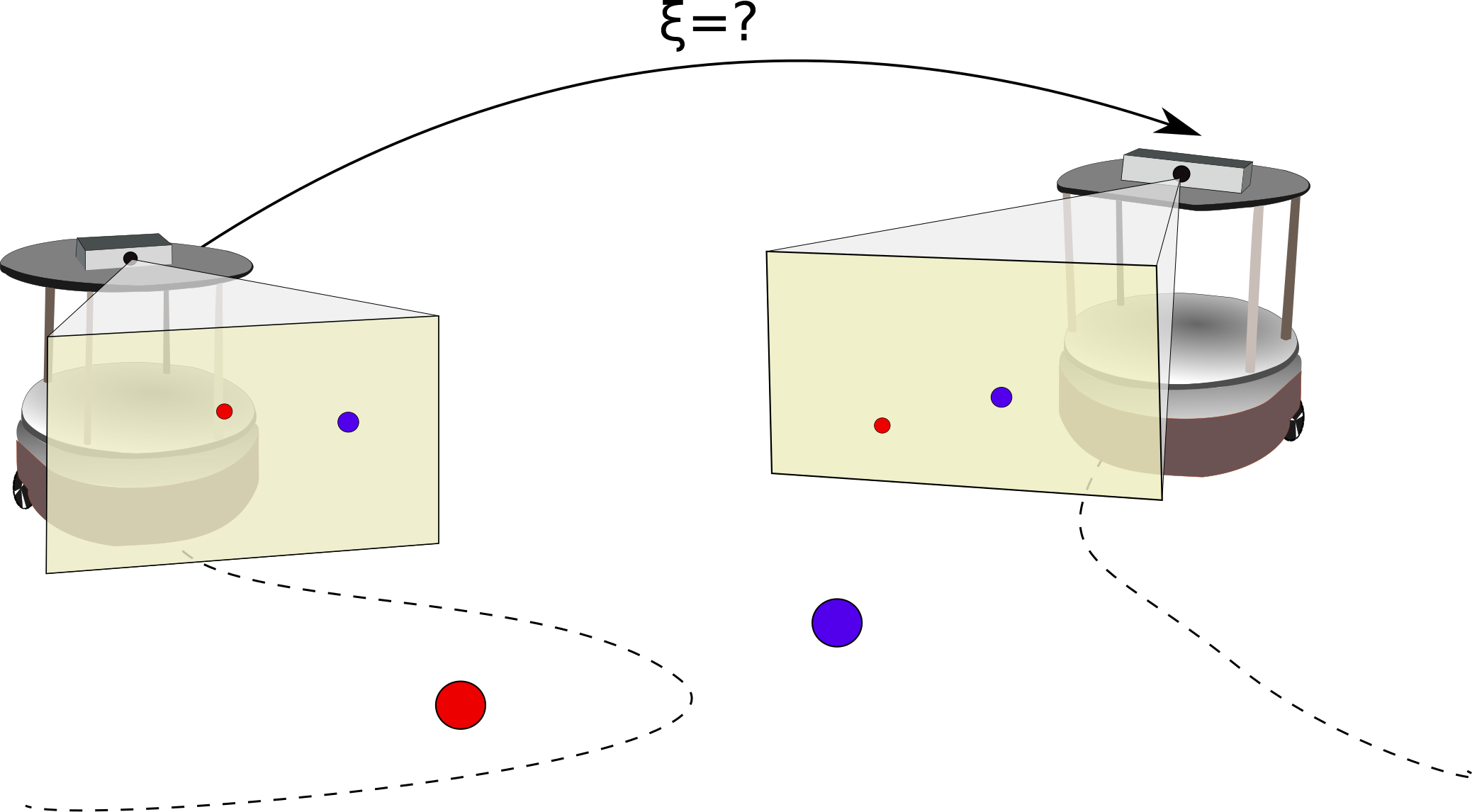}
    \caption{{\color{black}Representation of the problem addressed in this paper. Two robots are observing a pair of 3D points. We propose a pipeline to simultaneously estimate the 3D depth of the two 3D points, while at the same time, obtain their relative pose.}}
    \label{fig:intro}
\end{figure}

The decentralized coordinated control of multiple mobile robots has several applications, including exploration \cite{fox00}, coverage \cite{cortes04}, sampling and patrolling \cite{marino15}, search \& rescue \cite{jennings97}, and manipulation \cite{rus95}. Achieving coordination within a decentralized architecture requires each agent in the formation to have some knowledge of the pose of its neighboring agents. The relative pose information allows to locally optimize a goal function, re-plan trajectories, and avoid collisions. However, for some mission scenarios such as mine caves or catastrophic areas, it is particularly challenging to obtain an accurate estimate of the relative pose due to the constrained communication channel bandwidth and impossibility to install or employ a high-precision positioning system, e.g., geolocation devices (GPS and Galileo) or Motion Capture systems. 
This paper aims at studying the problem of relative localization for two autonomous non-holonomic ground robots using 3D points observed by an onboard visual system (See Fig.~\ref{fig:intro}). Since cameras observe 3D points up to an unknown scale, the formulation considers the problem of concurrently estimating the unknown depth of the 3D points. 

In monocular computer vision, the mapping task aims at estimating the inherent 3D structure of the world from the 2D pixels in the image frame \cite{hartley04}. Considering two or more sequences of images taking from different views in a static environment, the depth of the points is computed by estimating the 3D points where the correspondent projection rays intersect. More sophisticated solutions have been proposed \cite{artal15,oleynikova17,engel18}. However, these works do not take into account the motion imparted to the camera. Instead it is considered a sequence of images captured at a known pose in the environment. Some authors study the estimation of depth maps from a single image (e.g. \cite{choy16,fan17,huang18}). However, these works assume some high level knowledge of the environment. Recently, continuous 3D depth estimation has gained considerable attention (e.g. \cite{spica13,spica14,spica15,mateus18,romulo18}). The goal is to minimize the estimation error of the 3D features over a continuous sequence of images ordered in time, by accounting for the dynamics of the camera. In contrast to other solutions that aim at algebraically estimating the unknown parameters as a fitting problem, continuous estimation considers the error dynamics and its convergence to the origin over time. This strategy benefits from: 1) being able to provide a metric about the quality of their estimates, e.g. by measuring the error's covariance or convergence rate; 
and 2) having robust performance by continuously improving the estimation based on new views. These advantages are particularly important for image based visual servoing schemes \cite{chaumette06,spica17}, in which the depth of the 3D features are fed in the control loop.

Relative pose estimation is another active topic in the robotics and computer vision communities. From \cite{nister04}, geometrically one can see that the minimum required number of 2D point correspondences is five. The number of required point correspondences is reduced even more if the motion is constrained in a plane. In this case, only two 2D point correspondences are required \cite{ortin01}. One of the disadvantages of these methods is that they estimate the translation parameters up to a scale. This means that one needs to get additional information about the observed environment to get the complete relative pose.
Over the years, there have been many approaches to solve the relative pose problem, such as \cite{scaramuzza11,kneip14,ranftl18}. However, all of these methods were built on the assumption that the data (2D images) were discretely acquired over time, without any continuously time assumption. In \cite{spica16}, the authors propose a continuous perception strategy to estimate the relative localization of a set of robots in formation. Their method is based on the bearing-rigidity of the formation, and requires that the at least a pair robots should be in mutual visibility. In our work, we also aim at estimating the relative pose of the robots as in \cite{spica16}. 


The scenario considered in this paper includes two mobile robots, navigating autonomously in an unknown environment. Both robots are equipped with perspective cameras. 
While previous solutions for this problem consider a set of two or more images from the environment or use some special fleet configuration (e.g. the robots are in each others' fields of view or have the ability of sensing the bearing-information about each others' positions), in this paper we propose a framework that shares a set of common observations of the environment in the respective local frame of each robot (3D point features are employed). A graphical representation of the problem is shown in Fig.~\ref{fig:intro}. In this scenario, as will be shown in the experimental results, the proposed solution is able to estimate the 3 degrees of freedom (DoF) relative pose between the two robots using 3D points which are being continuously estimated by the two robotic agents.

To compute the relative pose, we use the fact that the two agents are tracking in the image plane two point correspondences. The proposed pipeline estimates the depth of the points seen by both robots while computing their relative pose. In contrast to \cite{ortin01}, by taking account a continuous estimation of the relative transformation, our method is able to get the complete relative pose (the translation is not estimated up to a scale). As the robots navigate in the environment, they will be estimating the depth of the points being tracked in their respective image frames using a depth estimation filter. Then, an Extended Kalman Filter (EKF) receives the estimated depths to estimate the relative pose of the robots.

This paper is organized as follows. The next section presents an overview of the system proposed in this paper. In Sec.~\ref{sec:ours} we describe the proposed method for depth and relative pose estimation. Results with simulated and real data are shown in Sec.~\ref{sec:results}, and in Sec.~\ref{sec:conclusions} we discuss and conclude the paper.
\section{SYSTEM OVERVIEW}
\label{sec:system}
This section presents the problem studied in this paper, an overview of the proposed system architecture, and the basic mathematical notation that holds for the remainder.

\begin{figure}[t]
    \centering
    \includegraphics[width=.45\textwidth]{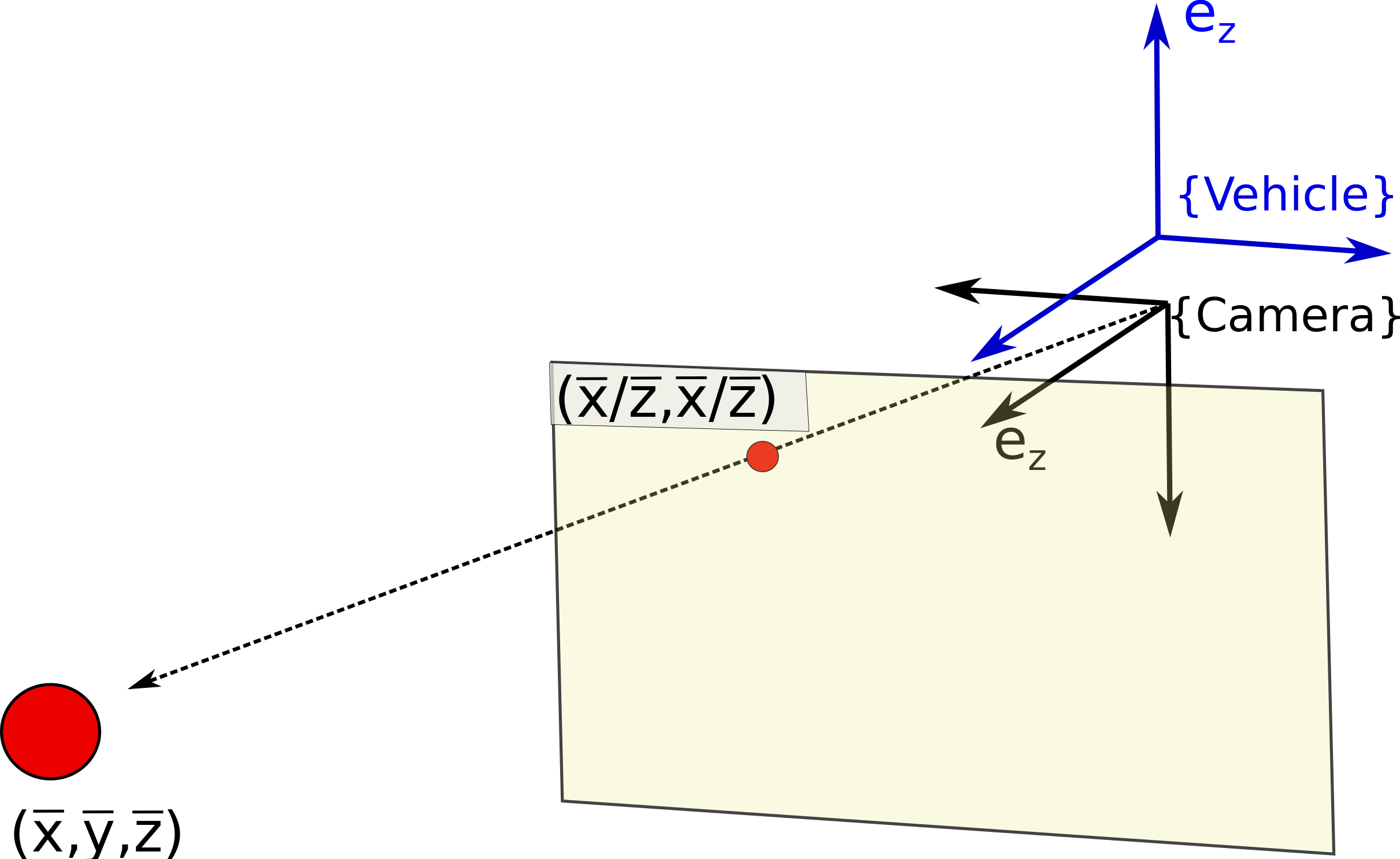}
    \caption{Configuration of the body-fixed and camera frames, and the notations used for the projection model.}
    \label{fig:frames}
\end{figure}

\begin{figure}[t]
    \centering
    \includegraphics[width=.45\textwidth]{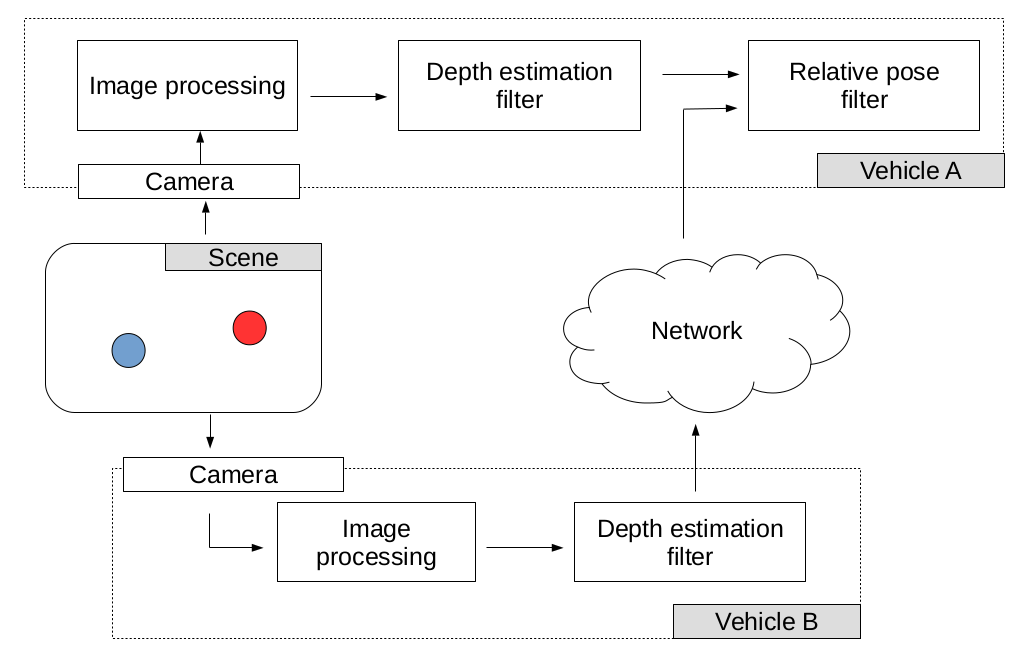}
    \caption{Pipeline of the solution proposed in this paper. Each robot is equipped with an RGB camera that captures images of two 3D points. Then, each robot has a depth estimation filter that estimates the depth of the observed points. All the estimates are shared through the network (including the robot's control input), and a filter estimates the relative pose between the robots.}
    \label{fig:pipeline}
\end{figure}

The problem discussed here is two-fold: 1) each agent has to compute the unknown depth of the points to recover the 3D structure of the world; and 2) each agent has to estimate the relative pose of its neighboring agent. The robot and camera configuration is shown in Fig.~\ref{fig:frames}, and the proposed pipeline solution is shown in Fig.~\ref{fig:pipeline}.

The cameras are mounted at the front of a non-holonomic vehicle, in which the optical axis of the sensor is aligned with the forward moving direction (see Fig.~\ref{fig:frames}). Without loss of generality, we assume that the camera lies in the origin of the vehicle. The robot actuation, denoted as $\mathbf{u}=[v_d, w_{\theta}]^T \in \mathbb{R}^2$, consists of a forward and an angular speed command. This translates into a linear speed along the optical axis of the camera and an angular speed around the $y$-axis of the camera.

A depth estimation filter runs in each agent independently, recovering the depth of the 3D points. The depth filter, discussed in Sec.~\ref{subsec:depth_filter}, is an particularization of \cite{spica14}, constrained for cameras mounted on non-holonomic planar vehicles. Then, an Extended Kalman Filter (EKF) gathers the motion of each camera and the coordinates of the observed points to estimate the relative pose between the vehicles. This requires that one vehicle sends its 3D points estimation and actuation commands to its neighboring vehicle through the network. This is discussed in Sec.~\ref{subsec:relative_pose_filter}. The EKF allows to capture the uncertainty inherent to the problem: depths converge online and measurements and actuation commands are subjected to noise.



The proposed pipeline includes two additional modules: image processing and network. The former is related with the detection and tracking of a 3D target on the image plane. For that purpose, one can use techniques such as SIFT~\cite{lowe99}, SURF~\cite{bay08}, or ORB~\cite{bradski11}. As a proof of concept, a color scheme detection is employed. It also eases the match of the correspondences between the points detected by the two robots. The network module enable the robots to communicate with each other. In practice, the vehicles are equipped with WiFi.

The mathematical notation adopted in this document is as follows: vectors are written in lower case bold font; matrices are written in upper case font; coordinate frames are denoted within curly brackets; and a subscript indicates the coordinate frame that a vector is described in (e.g. $\mathbf{u}_A$ stands for the control input of camera $A$).
\section{METHODS}\label{sec:ours}
This section presents the methods proposed for the simultaneous depth and relative pose localization of ground robots. We start by describing the used depth filter (Sec.~\ref{subsec:depth_filter}) for a single vehicle/point. Then, the relative pose filter is presented (Sec.~\ref{subsec:relative_pose_filter}). 

\subsection{Depth Filter}
\label{subsec:depth_filter}
This method is based on the framework proposed in \cite{spica13}, adapted to the motion of ground vehicles. Consider a camera moving freely in the 3D space, the dynamic of a point $\mathbf{p}=[\bar{x},\bar{y},\bar{z}]^T$ relative to the camera frame is
\begin{equation*}
    \dot{\mathbf{p}} = -\mathbf{w}\times\mathbf{p}-\mathbf{v} = \left\{\begin{matrix}
\dot{\bar{x}} = \bar{y}w_z - \bar{z}w_y - v_x\\ 
\dot{\bar{y}} = \bar{z}w_x - \bar{x}w_z - v_y\\
\dot{\bar{z}} = \bar{x}w_y - \bar{y}w_x - v_z
\end{matrix}\right., 
\end{equation*}
where $\mathbf{v}=[v_x,v_y,v_z]^T \in \mathbb{R}^3$ and $\mathbf{w}=[w_x,w_y,w_z]^T \in \mathbb{R}^3$ are the camera linear and angular velocity described in the camera frame. Since only the direction of the point is known, it is common practice to use the normalized image coordinates $\mathbf{s} = [\bar{x}/\bar{z},\bar{y}/\bar{z}]^T = [x,y]^T $. The previous equation can be re-stated as
\begin{equation*}
    \begin{bmatrix}
        \dot{x} \\ \dot{y}  
    \end{bmatrix} =
    \begin{bmatrix}
        -\frac{1}{\bar{z}} & 0 & \frac{x}{\bar{z}} & xy & -(1+x^2) & y \\
        0 & -\frac{1}{\bar{z}} & \frac{y}{\bar{z}} & 1+y^2 & -xy & -x \\
    \end{bmatrix}
    \begin{bmatrix}
    \mathbf{v} \\ \mathbf{w}
    \end{bmatrix}.
\end{equation*}

As discussed in Sec.~\ref{sec:system}, the vehicle input $\mathbf{u}=[v_d, w_\theta]^T$ consists in a forward speed ($v_d$) and an angular rate ($w_\theta$). {\color{black}The forward direction of the vehicle is aligned with the optical axis of the camera, i.e., the linear velocity expressed in the camera frame satisfies $v_x=v_y=0$ and $v_z = v_d$. The rotational axis of the vehicle is aligned with the y--axis of the camera such that $w_x=w_z=0$ and $w_y=-w_{\theta}$}. The reader is referred to Fig.~\ref{fig:frames} for a graphical illustration of the coordination frames at stake. Substituting $\chi=1/\bar{z}$ to obtain a linear expression and simplifying the previous equation by removing the components multiplied by $0$, the following dynamic system is obtained: 
\begin{align*}
    \dot{\mathbf{s}} & = f_m(\mathbf{s}, \mathbf{u}, t) + \Omega^T(t)\chi \\
    \dot{\chi} & = f_u(\mathbf{s}, \mathbf{u}, \chi, t),
\end{align*}
where 
\begin{equation}
\left\{\begin{matrix}
f_m(\mathbf{s}, \mathbf{u}, t) &=& w_{\theta} \begin{bmatrix}
    (1+x^2) \\ xy 
\end{bmatrix}\\ 
\Omega(t) &=& v_d\begin{bmatrix}
    x & y
\end{bmatrix} 
\\ 
f_u(\mathbf{s}, \mathbf{u}, \chi, t) &=& v_d\chi^2 + xw_{\theta}\chi
\end{matrix}\right. .
\label{eq:depth_filter_functions_3}
\end{equation}

In order to estimate the unknown depth $\bar{z}$, the estimation variables $\hat{\mathbf{s}}$ and $\hat{\chi}$ are introduced, as well as the associated estimation errors  $\tilde{\bm{s}} = \mathbf{s}-\hat{\mathbf{s}}$ and $\tilde{\chi} = {\chi}-\hat{{\chi}}$. For that purpose, the following observer is considered:
\begin{align}
\dot{\hat{\mathbf{s}}} &= f_m(\mathbf{s}, \mathbf{u}, t) + \Omega^T(t)\hat{\chi} + H\tilde{\bm{s}} 
\label{eq:observer_s}\\
\dot{\hat{{\chi}}} &= f_u(\bm{s}, \mathbf{u},  \chi, t) + \Lambda\Omega^T(t)Q\tilde{\bm{s}},
\label{eq:observer_chi}
\end{align}
where $\Lambda \in \mathbb{R}_{++}$ is a positive gain and $H, Q = \alpha I_2 \in \mathbb{S}^2$ are symmetric positive matrices with $\alpha > 0$. Since for non-null $v_d$ the camera moves along the optical axis, $\dot{\chi} \neq 0$. For this setup, the dynamics of the error can be proven to be locally exponentially stable. For a formal proof of the convergence of error to the origin, the reader is referred to \cite{spica13}. The basic idea consists in guarantee that a Persistency of Excitation conditions holds. More precisely, this requires that the matrix $\Omega(t)\Omega(t)^T$ is full rank throughout time, that is
\begin{equation*}
    \int_t^{t+T} \Omega(\tau)\Omega^T(\tau)d\tau \geq \gamma I > 0, \quad t \geq 0,
\end{equation*}
where $\gamma > 0$ and $I$ is an identity matrix of appropriate dimension. In particular, for a single point, this can be done by ensuring that $\|\Omega(t)\|^2 > 0$. From \eqref{eq:depth_filter_functions_3}, we have
\begin{equation*}
\|\Omega(t)\|^2 = (x^2 + y^2) v_d^2.
\end{equation*}
{\color{black}Therefore, the filter converges, as long as the projected point is not in the center of the image frame, i.e., $x\neq 0$ or $y\neq0$ must hold for convergence. Geometrically, this constraint is easy to verify: if $x=y=0$ and the camera moves along the $z$--axis, the motion will be along along the projection ray meaning and it would be impossible to recover the depth (not enough information to do triangulation)}.

In the next section, the recovered 3D points are employed for online relative pose estimation.

\subsection{Relative Pose Filter}
\label{subsec:relative_pose_filter}
This section presents relative pose filter, which takes as input the depth estimation and the control inputs of the robots. First, consider that each robot has a body-fixed coordinate frame, denoted as $\{A\}$ for robot $A$ and $\{B\}$ for robot $B$. The transformation from the body-fixed to the camera fixed frame is known for each robot. Now, instead of a single point, the robots observe $m$--points in the scenario. For each observed 3D point, each robot runs an instance of the depth estimation filter, as presented before. The pose of $B$ with respect to $A$ described in $\{A\}$ is denoted as $\bm{\xi} = [\mathbf{r}^T, \theta]^T \in \mathit{SE}(2)$, where $\mathbf{r} = [r_x,r_y]^T$ is a 2D translation and $\theta$ an orientation angle. In accordance with the definition presented in Sec. \ref{sec:system}, the control input for each vehicle is defined as $\mathbf{u}_A = [v_{A,d}, w_{A,\theta}]^T$ and $\mathbf{u}_B = [v_{d,B}, w_{B,\theta}]^T$. The list of notations is summarized in  Table~\ref{table:list_of_variables}.   

\begin{table}[t]
\caption{List of Notations for the relative pose estimation filter.}
\label{table:list_of_variables}
\centering
\begin{tabular}{c|l}
\textbf{Variable}            & \textbf{Description} \\ \hline
$A$, $B$ & Vehicles \\
$\{A\}$, $\{B\}$ & Body-fixed coordinate frames \\
$\mathbf{u}_A$, $\mathbf{u}_B$ & Control input of camera A and B \\
$(\mathbf{s}^A_i)_{i=1}^m$, $(\mathbf{s}^B_i)_{i=1}^m$ & Normalized coordinates of the 3D points \\
$(Z_i^A)_{i=1}^m$, $(Z_i^B)_{i=1}^m$ & Depth of the 3D points \\
$\bm{\xi}$ & Pose of $B$ w.r.t $A$ described in $\{A\}$ \\
\end{tabular}
\end{table}

In order to derive the dynamic model of the relative pose between the robots, let $\bm{\xi}_A=[\mathbf{r}_{A}^T, \theta_{A}]^T$ and $\bm{\xi}_B=[\mathbf{r}_{B}^T, \theta_{B}]^T$ be the pose of robot $A$ and $B$, respectively, described in an inertial coordinate frame. The dynamic of each robot is
\begin{equation}
\left\{\begin{aligned} 
\dot{r}_{A,x} & = v_{A,d}\cos{\theta_A}  \\ 
\dot{r}_{A,x} & = v_{A,d}\sin{\theta_A}  \\ 
\dot{\theta}_{A} & = w_{A,\theta}
\end{aligned}\right.
\, \text{ and } \,
\left\{\begin{aligned}
\dot{r}_{B,x} & = v_{B,d}\cos{\theta_B}  \\
\dot{r}_{B,y} & = v_{B,d}\sin{\theta_B}  \\ 
\dot{\theta}_{B} & = w_{B,\theta}
\end{aligned}\right. .
\label{eq:pose_inertial}
\end{equation}
The relative pose between  the two non-holonomic vehicles described in $\{A\}$ is given by
\begin{equation}
\bm{\xi} = \begin{bmatrix} \mathbf{r} \\ \theta \end{bmatrix}=\begin{bmatrix}
    R^T(\theta_A)(\mathbf{r}_B - \mathbf{r}_A) \\
    \theta_B - \theta_A
\end{bmatrix},
\label{eq:rel_pose}
\end{equation}
where the rotation matrix $R(\cdot)$ will be soon defined. The dynamics of the relative pose is
\begin{equation}
\dot{\bm{\xi}} = \begin{bmatrix}
    \dot{R}^T(\theta_A)(\mathbf{r}_B - \mathbf{r}_A) + {R}^T(\theta_A)(\dot{\mathbf{r}}_B - \dot{\mathbf{r}}_A) \\
    \dot{\theta}_B - \dot{\theta}_A
\end{bmatrix},
\label{eq:dynamic_01}
\end{equation}
where $\dot{R}(a) = R(a)S(\dot{a})$,
\begin{equation}
R(a) = \begin{bmatrix}
    \cos a & -\sin a\\
    \sin a & \cos a    
\end{bmatrix}, \text{ and } 
S(\dot{a}) = \begin{bmatrix}
    0 & -\dot{a}\\
    \dot{a} & 0    
\end{bmatrix}.
\label{eq:matrices_definition}
\end{equation}
Substituting \eqref{eq:rel_pose} in \eqref{eq:dynamic_01} yields 
\begin{equation}
\dot{\bm{\xi}} = \begin{bmatrix}
    -S(\dot{\theta}_A)\mathbf{r}  + {R}^T(\theta_A)(\dot{\mathbf{r}}_B - \dot{\mathbf{r}}_A) \\
    \dot{\theta}_B - \dot{\theta}_A
\end{bmatrix}.
\label{eq:rotation_matrix}
\end{equation}
Finally, applying \eqref{eq:pose_inertial} and \eqref{eq:matrices_definition} in the previous equation leads to the dynamic model
\begin{equation}
 \dot{\bm{\xi}} = f(\bm{\xi}, \mathbf{u}_A, \mathbf{u}_B) =
\begin{bmatrix}
    v_{B,d}\cos{\theta} - v_{A,d} + {w}_{A, \theta}r_y\\
    v_{B,d}\sin{\theta} -{w}_{A, \theta}r_x \\
    {w}_{B,\theta} - {w}_{A, \theta} 
\end{bmatrix}.
\label{eq:state_model}
\end{equation}

The output model considers the relationship between the pose of the robots and the observed 3D points in the body-fixed frame. From the geometric perspective, the following equation holds for each of the $m$--points: 
\begin{equation*}
  \mathbf{r} = {\bar{z}^A_i\mathbf{s}^{A}_i} - R(\theta){\bar{z}^B_i\mathbf{s}^{B}_{i}},
\end{equation*}
for which the corresponding time derivative is
{\small
\begin{equation*}
    \dot{\mathbf{r}} = \dot{\bar{z}}_i^A\mathbf{s}_i^A + \bar{z}_i^A\dot{\mathbf{s}}_i^A - R(\theta)[S(\dot{\theta})\bar{z}^B_i\mathbf{s}^B_i + \dot{\bar{z}}_i^B\mathbf{s}_i^B + \bar{z}_i^B\dot{\mathbf{s}}_i^B].
\end{equation*}
}By replacing $\dot{\bar{z}} = -\dot{\chi}/{\chi}^2$, it is possible to describe the relationship between the dynamics of 3D points, stated in \eqref{eq:observer_s} and \eqref{eq:observer_chi}, and the dynamics of the relative pose:
\begin{equation*}
\footnotesize
 \dot{\mathbf{r}}  =-\frac{\dot{\chi}_i^A}{(\chi_i^A)^2}\mathbf{s}_i^A + \frac{\dot{\mathbf{s}}_i^A}{\chi_i^A} - {R}(\theta)\left[S(\dot{\theta})\frac{\mathbf{s}^B_i}{\chi^B_i} - \frac{\dot{\chi}_i^B}{(\chi_i^B)^2}\mathbf{s}_i^B + \frac{\dot{\mathbf{s}}_i^B}{\chi_i^B}\right].
\end{equation*}

{\color{black}If vehicle $A$ periodically estimates $(\mathbf{s}_i^A, {\chi}_i^A)$ and their dynamics, and receives the corresponding estimates from vehicle $B$, the following observation model holds}:
\begin{equation*}
  h_i(\bm{\xi}) = R^T(\theta)[\dot{\bar{z}}_i^A\mathbf{s}_i^A + \bar{z}_i^A\dot{\mathbf{s}}_i^A].
\end{equation*}

The relative pose filter consists in an Extended Kalman Filter that assumes that the input of the vehicles and the observations are perturbed by Gaussian noise with zero mean. First, in the prediction step, the prior is obtained according to  \eqref{eq:state_model}
\begin{equation*}
    \dot{\hat{\bm{\xi}}}^- = f(\dot{\hat{\bm{\xi}}}, \mathbf{u}^A, \mathbf{u}^B)
\end{equation*}
Then, for each pair of measurements, the update model is:
\begin{equation*}
\dot{\hat{\bm{\xi}}} = \dot{\hat{\bm{\xi}}}^- + K[\pi_i(t) - h_i(\dot{\hat{\bm{\xi}}}^-)],
\end{equation*}
where K is the Kalman gain and 
\begin{equation*}
\pi_i(t) = S(\dot{\theta})\bar{z}^B_i\mathbf{s}^B_i + \dot{\bar{z}}_i^B\mathbf{s}_i^B + \bar{z}_i^B\dot{\mathbf{s}}_i^B
\end{equation*}
is computed from the measurements received from the neighboring vehicle. The Kalman gain and the covariance matrix are updated following the regular EKF framework. 


With this we conclude the relative pose filter description. This setup allows us to obtain an estimate of the unknown state. We stress that this method builds on the assumption that at least two 3D points correspondences are required to align two coordinate systems \cite{miraldo19}. Although we focus on the critical case with only two points, in general, the EKF framework benefits from multiple observations and increase robustness against occlusions. 


\section{Results}\label{sec:results}
In this section, we discuss simulated (Sec. ~\ref{sec:exp_simulated} ) and experimental results (\ref{sec:exp_real}). Simulations were performed using the Robot Operating System (ROS) and Gazebo/Turtlebot simulator at $20$ [Hz]. Similarly, for the experimental results, ROS and two TurtleBots were employed. A footage of this experiment is attached.  

\subsection{Simulation}\label{sec:exp_simulated}
\begin{figure}
    \centering
    \includegraphics[width=.475\textwidth]{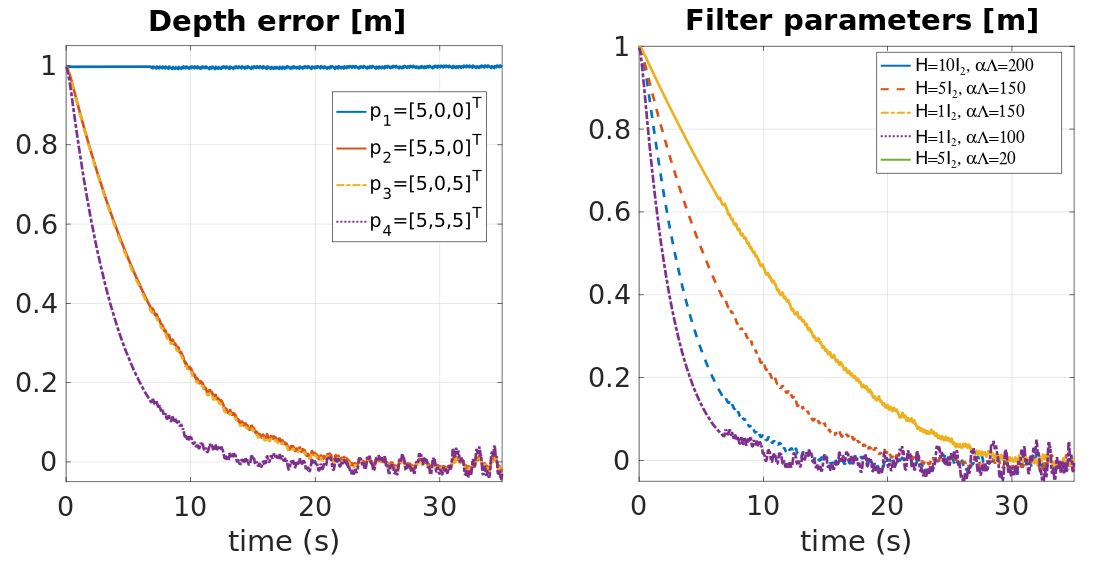}
    \caption{Depth estimation for different scenarios using Gazebo/Turtlebot. Robot moves forward and does not rotate. At the left we show the depth estimation error for 4 different points. The graphic at the right shows the convergence behaviour for a single point and different filter parameters.}
    \label{fig:depth_estimation_analysis.}
\end{figure}

The first simulated scenario illustrates the depth filter properties presented in Sec.~\ref{subsec:depth_filter}. A single non-holonomic ground robot observes four 3D points. At $t=0$, the points coordinates w.r.t. to the camera frame are $\mathbf{p}_1=[0,0,5]^T$, $\mathbf{p}_2=[0,5,5]^T$, $\mathbf{p}_3=[5,0,5]^T$, and $\mathbf{p}_4=[5,5,5]^T$. For each point, a filter was initialized with 1 [m] depth estimation error. The robot actuation is constant, set to $v_d=$ 0.1 [m/s] and $w_\theta=$ 0 [rad], that is, the robot moves straight ahead in the direction of the optical axis. Figure~\ref{fig:depth_estimation_analysis.} shows the results for the three points using the same filter parameters. The estimation error converges to zero when the point does not rest at the origin of the image frame. 
\begin{figure}
    \centering
    \includegraphics[width=.475\textwidth]{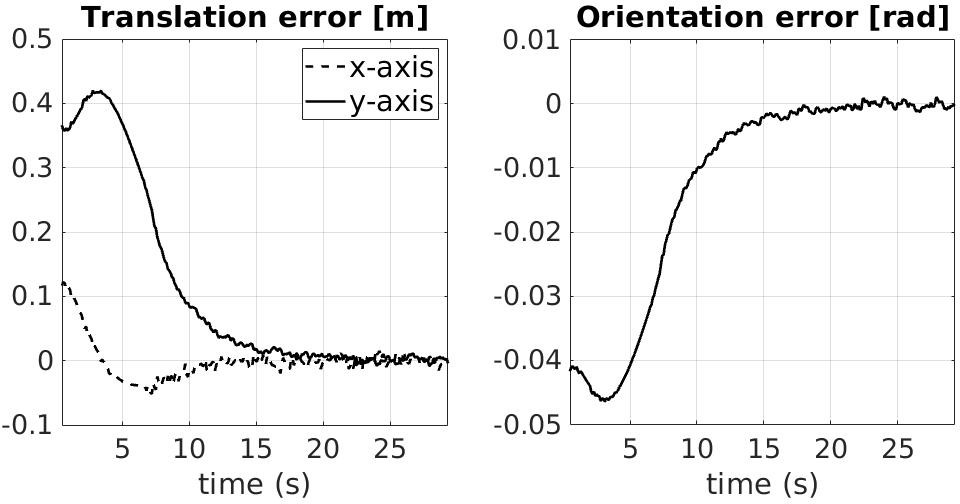}
    \caption{Relative pose estimation: (a) translation and (b) orientation. The initial estimation error is 1.5 m in the $x$,$y$ axes and 15 deg. in orientation.}
    \label{fig:gazebo_rel_pose.}
\end{figure}
In fact, the further the point is from the center of the image, the faster is the convergence due to the larger value of $\Omega(t)$. However, this also amplifies numerical errors and noise. Figure~\ref{fig:depth_estimation_analysis.} at the right depicts the impact of different gains on the depth estimation, $\mathbf{p}_3$ was selected. Larger values of $H$ in \eqref{eq:observer_s} slow down the convergence rate, while higher values  of $\alpha\Lambda$ increase the convergence rate. Here, once again, the designer must take into consideration that faster convergence leads to large steady-state errors. 

Figure~\ref{fig:gazebo_rel_pose.} shows the performance of the proposed depth estimation and relative pose estimation framework (method proposed in Sec.~\ref{subsec:relative_pose_filter}) altogether. Only the points $\mathbf{p}_1$ and $\mathbf{p}_4$ from the previous experience were considered and an additional robot was inserted in the scenario. For each set of robot/point, the initial depth estimation error was set to $1$ [m] and the initial relative pose estimation error was set to $1.5$ [m] in each transnational axis and $15^o$ in orientation. The depth estimation error for each point follows a convergence behaviour similar to the one shown in Fig.~\ref{fig:depth_estimation_analysis.}. Therefore, as shown in Fig.~\ref{fig:gazebo_rel_pose.}, the translation estimation error converges to zero as the depth estimation error converges to zero. The same applies for the orientation error.

\subsection{Experimental Results}\label{sec:exp_real}
\begin{figure}[t]
    \centering
    \includegraphics[width=0.405\textwidth]{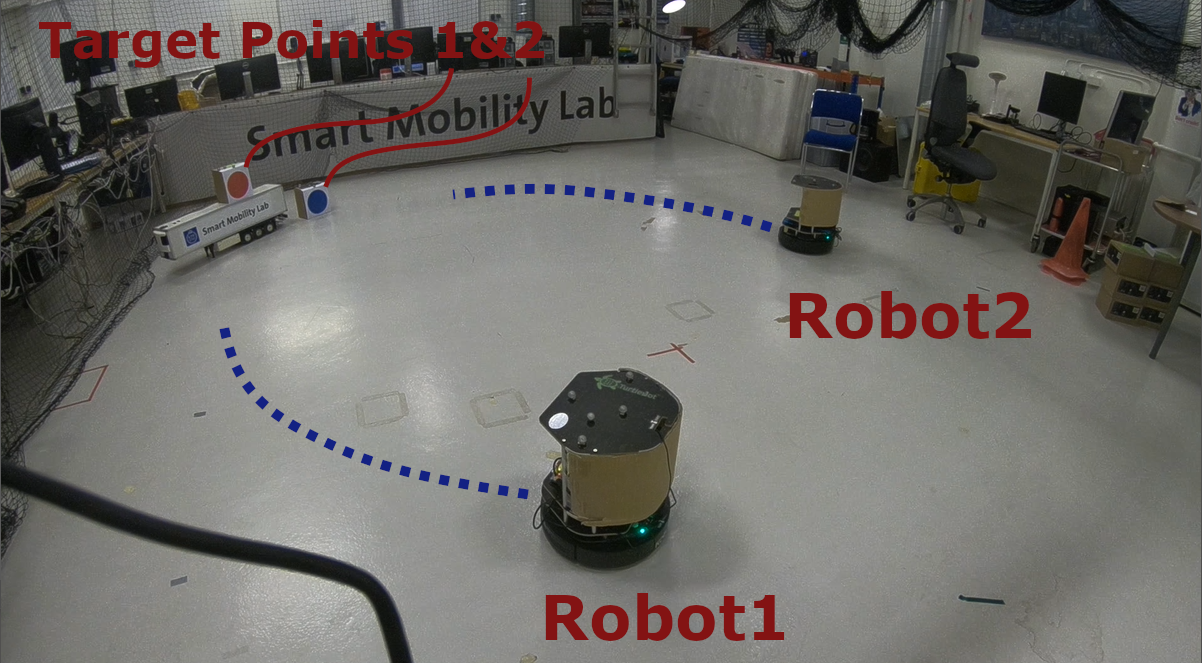}
    \caption{Setup used in the experimental results. Two mobile robots are navigating in the environment while observing two 3D points (the red and blue targets).}
    \label{fig:exp_setup}
\end{figure}

{\color{black}Experiments were conducted at the Smart Mobility Lab\footnote{https://www.kth.se/dcs/research/control-of-transport/smart-mobility-lab/smart-mobility-lab-1.441539 [accessed \today].}, from KTH. We employed two TurtleBots (differential drive robots) named here {\tt Robot\_A} and {\tt Robot\_B}, respectively. Each robot was equipped with a perspective camera pointing towards the motion of the robot. The scenario contained two 3D targets (red and blue targets). For evaluation purpose, a motion capture system installed at the ceiling of the arena provides ground truth for the poses of the robots and the 3D points.} Fig.~\ref{fig:exp_setup} shows the proposed setup. The following velocities were sent to the robots: $v_d^A$ = 0.1[m/s] \& $w_\theta^A$ = 0.02[rad/s], and $v_d^B$ = 0.1[m/s] \& $w_\theta^B$ = -0.03[rad/s], for {\tt Robot\_A} and {\tt Robot\_B}, respectively. Through all the trajectory, we make ensure that the robots are observing the target points (around 40 [s]). The depth filter parameters were $H=2.5I$ and $\alpha\Lambda=120$ for both robots.

\begin{figure}
    \centering
    \includegraphics[width=.425\textwidth]{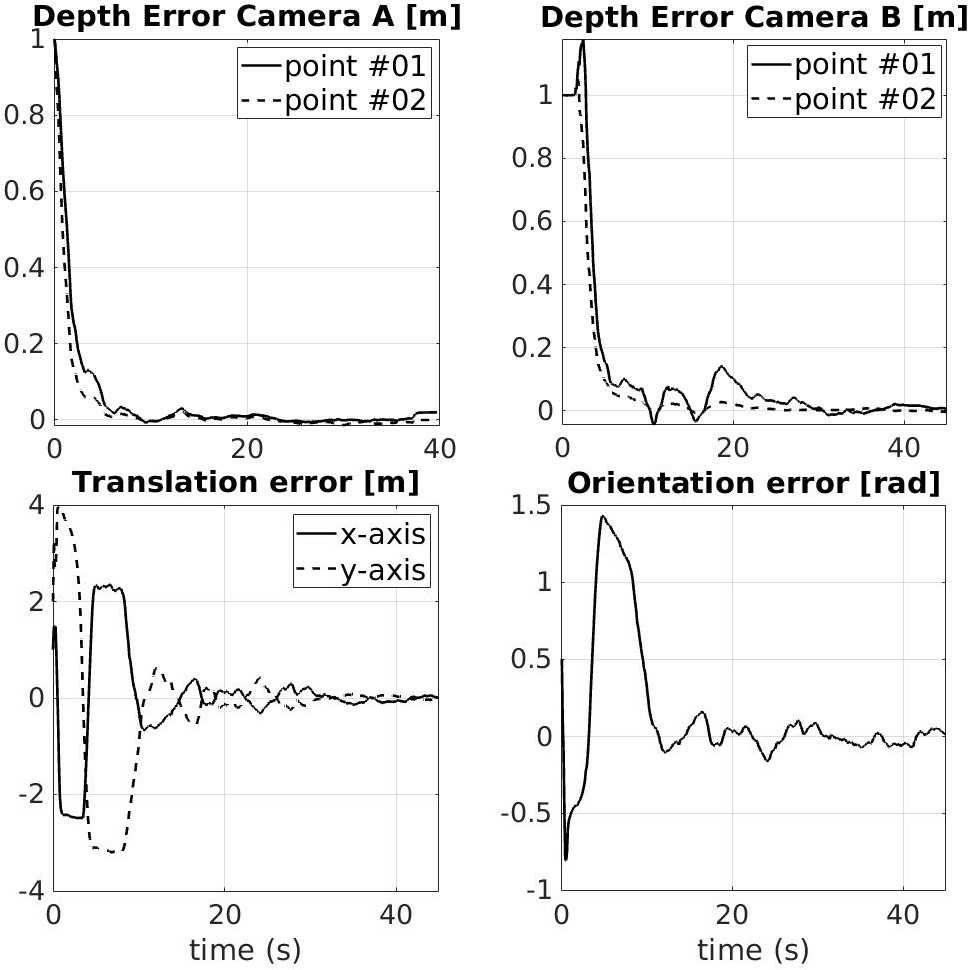}
    \caption{Experimental results in a scenario with two 3D points: (a) and (b) shows the depth estimation error of the 3D points converging to the origin for robots $A$ and $B$, respectively. Robot A estimates the pose of B described in $\{A\}$: (c) shows the relative translation estimation error and (d) the relative orientation estimation error.}
    \label{fig:experimental_results}
\end{figure}

The results are shown in Fig. \ref{fig:experimental_results}. The initial error for the depth filters were approximately 1 [m]. The depth of both points converge after $13$ [s] for {\tt Robot\_A} and $25$ [s] for {\tt Robot\_B}. The offset between the initial guess and the real values for the relative pose filter was 1.5 [m] in each of the translation axis and $60^o$ in orientation. As one can see, after a transient period, the errors converge to zero, proving that the proposed pipeline is able to correctly estimate the relative pose between the two robots for the evaluated setup.

\section{DISCUSSION}\label{sec:conclusions}
In this paper we presented a framework for depth and relative pose estimation for two vision-based robots. The depth estimation filter was built on top of recent results on continuous structure from motion methods. In particular, we constrained the problem to a camera mounted on a non-holonomic ground vehicle. The depth estimation and robot input commands are fed into an Extended Kalman Filter that estimates the relative pose between the ground robots. Simulations and experiments in a real scenario shown the convergence of both filters for several scenarios and set of initial conditions.
In future work, the active control of the robots will be considered as well as the interplay between the gains and the position of the point in the image frame. 

\section*{ACKNOWLEDGEMENT}
The authors are grateful for the invaluable support provided by Pedro Roque (KTH) in the experimental setup.


\bibliographystyle{IEEEtran}
\bibliography{files/sacrpe}

\end{document}